\documentclass[letterpaper, 10 pt, conference]{IEEEtran}
\IEEEoverridecommandlockouts

\usepackage{cite}
\usepackage{amsmath,amssymb,amsfonts}
\usepackage{algorithmic}
\usepackage{graphicx}
\usepackage{textcomp}
\usepackage{cite}
\usepackage{multirow}
\usepackage{hyperref}
\usepackage[font=small]{caption}
\usepackage{subcaption}

\usepackage[table,xcdraw]{xcolor}

\def\BibTeX{{\rm B\kern-.05em{\sc i\kern-.025em b}\kern-.08em
    T\kern-.1667em\lower.7ex\hbox{E}\kern-.125emX}}

\makeatletter
    \newcommand{\linebreakand}{%
      \end{@IEEEauthorhalign}
      \hfill\mbox{}\par
      \mbox{}\hfill\begin{@IEEEauthorhalign}
    }
\makeatother

\title{ViewVR: Visual Feedback Modes to Achieve Quality of VR-based Telemanipulation

\author{\IEEEauthorblockN{Artem Erkhov}
\IEEEauthorblockA{\textit{Skoltech} \\
Moscow, Russia \\
Artem.Erkhov@skoltech.ru}
\and
\IEEEauthorblockN{Artem Bazhenov}
\IEEEauthorblockA{\textit{Skoltech} \\
Moscow, Russia \\
Artem.Bazhenov@skoltech.ru}
\and
\IEEEauthorblockN{Sergei Satsevich}
\IEEEauthorblockA{\textit{Skoltech} \\
Moscow, Russia \\
Sergei.Satsevich@skoltech.ru}
\linebreakand
\IEEEauthorblockN{Danil Belov}
\IEEEauthorblockA{\textit{Skoltech} \\
Moscow, Russia \\
Danil.Belov@skoltech.ru}
\and
\IEEEauthorblockN{Farit Khabibullin}
\IEEEauthorblockA{\textit{Skoltech} \\
Moscow, Russia \\
Farit.Khabibullin@skoltech.ru}
\and
\IEEEauthorblockN{Sergei Egorov}
\IEEEauthorblockA{\textit{Skoltech} \\
Moscow, Russia \\
Sergei.Egorov@skoltech.ru}
\linebreakand
\IEEEauthorblockN{Maxim Gromakov}
\IEEEauthorblockA{\textit{Skoltech} \\
Moscow, Russia \\
Maxim.Gromakov@skoltech.ru}
\and
\IEEEauthorblockN{Miguel Altamirano Cabrera}
\IEEEauthorblockA{\textit{Skoltech} \\
Moscow, Russia \\
M.Altamirano@skoltech .ru}
\and
\IEEEauthorblockN{Dzmitry Tsetserukou}
\IEEEauthorblockA{\textit{Skoltech} \\
Moscow, Russia \\
D.Tsetserukou@skoltech.ru}
}}

\begin{document}

\maketitle

\begin{abstract}
The paper focuses on an immersive teleoperation system that enhances operator’s ability to actively perceive the robot’s surroundings. A consumer-grade HTC Vive VR system was used to synchronize the operator’s hand and head movements with a UR3 robot and a custom-built robotic head with two degrees of freedom (2-DoF). The system’s usability, manipulation efficiency, and intuitiveness of control were evaluated in comparison with static head camera positioning across three distinct tasks. Code and other supplementary materials can be accessed by link:  https://github.com/ErkhovArtem/ViewVR.
\end{abstract}




\section{Introduction}

Teleoperation plays a pivotal role in robotics by enabling efficient data collection for learning from demonstrations. The quality of collected data heavily depends on the operator’s ability to intuitively control the system and receive adaptive visual feedback. Traditional static camera systems often fail to provide sufficient flexibility and viewpoints necessary for complex manipulations. Recent advancements, such as camera-in-hand configurations and multi-camera setups, have aimed to address these limitations but can introduce challenges like increased system complexity and operator disorientation \cite{rakita2017motion, lin2024learning}.

To overcome these issues, we propose an immersive teleoperation system that integrates a consumer-grade HTC Vive VR setup and a 2-DoF actuated robotic head. This approach enhances visual feedback by aligning the robot's camera perspectives with the operator’s head movements, significantly improving spatial awareness and control precision. Drawing inspiration from systems like multi-DoF robotic head \cite{torso} and camera-in-hand configurations \cite{rakita2017motion}, our work emphasizes usability and manipulation efficiency in diverse tasks.

We evaluated our system against traditional fixed-camera setup across three tasks of varying difficulty, demonstrating its potential for improving task success rates and user experience. This work contributes to the advancement of teleoperation systems by offering a scalable solution that bridges the gap between intuitive control and adaptive feedback.

\begin{figure}[t]
    \centering
    \includegraphics[width=1\linewidth]{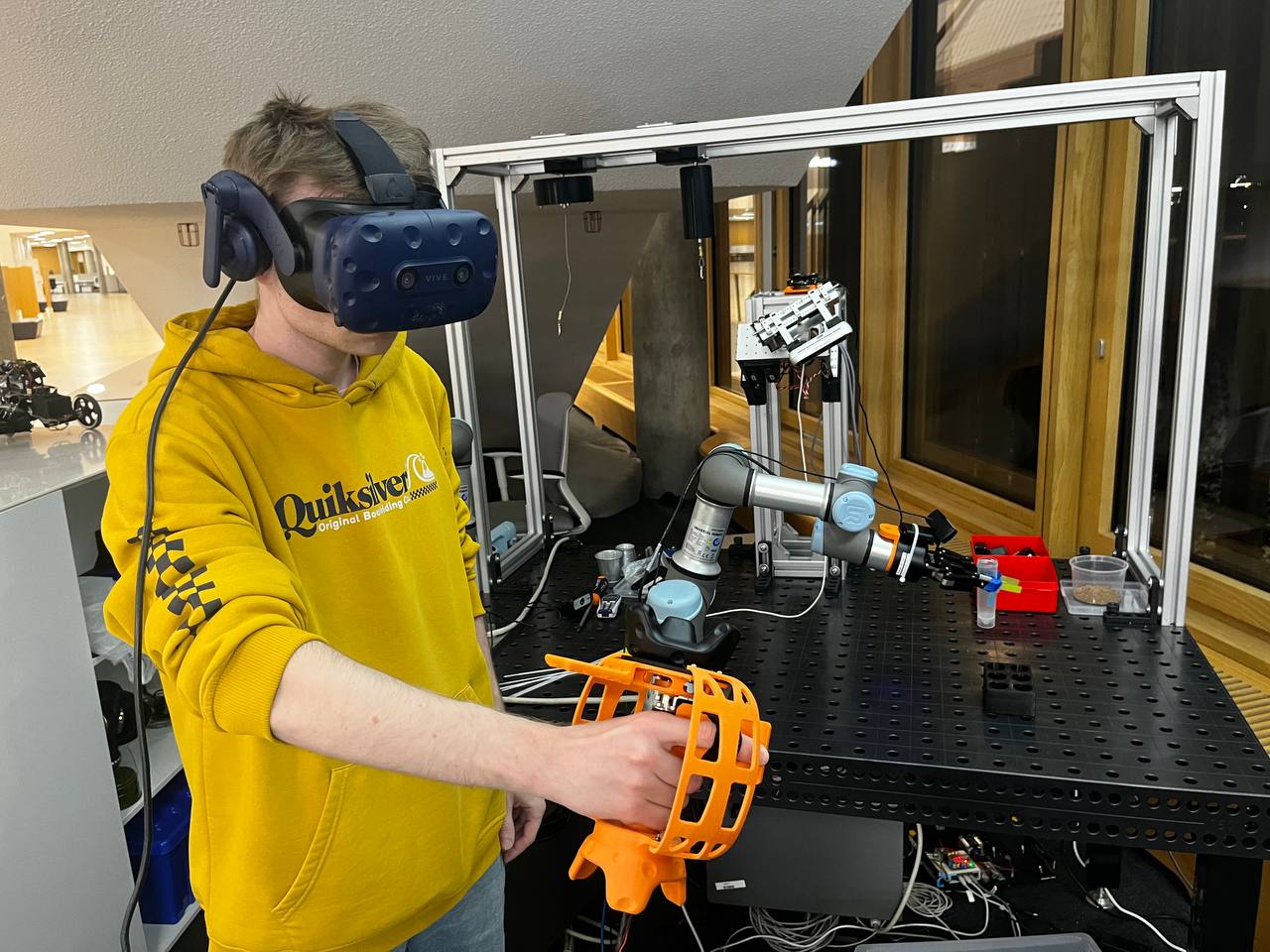}
    \caption{Operator teaches robot to manipulate the objects with ViewVR.}
    \label{fig:main}
    \vspace{-0.2cm}
\end{figure}

\section{Related Work}

\subsection{Advancements in Teleoperation}

The field of teleoperation has advanced significantly, with diverse camera configurations enhancing operator performance and task efficiency. Key contributions include dynamic camera systems, immersive VR environments, multi-camera setups, and user-centered frameworks, which enables an operator at the control to perform remote tasks dexterously with the feeling of existing in a surrogate robot working in a remote environment \cite{telexistence}.

Dynamic camera systems, such as the “camera-in-hand” approach \cite{rakita2017motion} and autonomous multi-camera systems \cite{kamezaki2016visibility}, adapt viewpoints in real time to reduce occlusions and improve task performance. Models for viewpoint adjustment \cite{zhu2011exploring} further support multi-tasking scenarios, addressing operator disorientation and limited visibility.

Immersive VR environments enhance engagement and precision by integrating VR headsets with robotic systems \cite{yim2022wfh} and employing intuitive interfaces \cite{barentine2021vr}. These systems reduce cognitive load and improve spatial awareness, while frameworks like \cite{gallipoli2024virtual} allow operators to switch between immersive and conventional modes for greater flexibility. Similarly, AeroVR integrates a VR-based teleoperation system for aerial manipulation, combining robotic arms with a digital twin environment to improve stability and usability during complex tasks \cite{8981574}.

Multi-camera setups, combining wrist-mounted and overhead views \cite{audonnet2024telesim}, enhance depth perception and spatial awareness. Dynamic camera fusion \cite{keyes2006camera} optimizes viewpoints for real-time tasks, underscoring the importance of tailored camera configurations in teleoperation. DronePick enables remote object picking and delivery using a quadcopter controlled by a tactile glove, providing both visual and haptic feedback for precision in operations \cite{ibrahimov2019dronepick}.

Shared autonomy frameworks balance human input with robot functions, streamlining tasks in dynamic settings \cite{manschitz2022shared}. User-centered approaches \cite{rea2022still} prioritize intuitive interfaces and visual feedback to mitigate latency and occlusions, aligning system design with operator needs.

Advanced visualization techniques, such as active stereo cameras mimicking head movements \cite{cheng2024open} and mixed-reality environments \cite{su2022mixed}, further expand teleoperation capabilities. GraspLook introduces an augmented virtual environment enhanced by a region-based convolutional neural network (R-CNN) for detecting and manipulating objects in medical and industrial settings. This system reduces the cognitive load and task execution time while increasing operator performance compared to camera-based teleoperation \cite{9659460}.

Applications span healthcare, industrial automation, and exploration. For instance, augmented virtuality systems with haptic feedback \cite{gonzalez2021advanced} improve industrial task execution, while motion retargeting methods \cite{rakita2017motion} enable intuitive robotic arm control.


\subsection{Existing Camera Configurations}

A simple and cost-effective approach to observing a manipulator's workspace is using stationary cameras, as demonstrated by Ding et al. \cite{ding2024bunny}. While straightforward, this method often lacks sufficient angles for detailed manipulations and may result in obstructed views.

To overcome these limitations, researchers have explored combining stationary and wrist-mounted cameras positioned near the robot’s end effector \cite{lin2024learning, li2024haptic, pan2024vision, fu2024learning, kamijo2024learning}. Systems employing multiple stationary cameras, such as those by Naceri et al. \cite{naceri2021vicarios}, and setups integrating top, bottom, and wrist-mounted views, as in ALOHA 2 \cite{aldaco2024aloha}, improve coverage. Zhao et al. \cite{zhao2023learning} add front and wrist-mounted cameras, while Wang et al. \cite{wang2023mimicplay} extend this with left- and right-corner views for greater flexibility.

For precise tasks, additional adjustments are often needed. Chuang et al. \cite{chuang2024active} address this by equipping a 7-DoF robotic arm with a stereo camera, enabling autonomous perspective optimization. Their study shows that such configurations enhance task execution and reduce reliance on specific camera setups. Similarly, Rakita et al. \cite{rakita2017motion} propose a second robotic arm with a camera-in-hand system, demonstrating superior performance in remote teleoperation scenarios.

An alternative approach involves mounting cameras on multi-DoF robotic heads to replicate operator head movements, as described by Cheng et al. \cite{cheng2024open}, Xu et al. \cite{xu2022shared}, and Zhang et al. \cite{zhang2018deep}. Cheng et al. report that this method improves view exploration, though usability and efficiency comparisons remain limited.

Ben et al. \cite{sen2024learning} extend this concept with cameras mounted on a 5-DoF robotic neck, torso, and wrists. Their system reduces cognitive load, enhances situational awareness, and improves efficiency in complex environments with visual occlusions.

\section{System Architecture}

We propose a modular teleoperation system based on the HTC Vive PRO VR setup and a 2-DoF robotic head equipped with an Intel RealSense D455 camera, designed for high-dexterity manipulation using the UR3 6-DoF robotic arm.

\begin{figure}[htp]
    \centering
    \includegraphics[width=8cm]{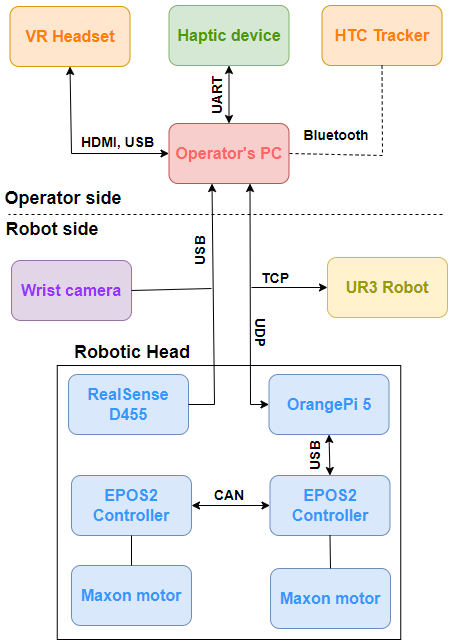}
    \caption{System architecture of ViewVR.}
    \label{fig:system architecture}
    \vspace{-0.3cm}
\end{figure}

Our system consists of two subsystems: the Operator (local) and Robot (remote) subsystems, connected via Ethernet (\autoref{fig:system architecture}). On the operator's side, we use the HTC Vive PRO headset and the HTC Tracker 2.0 to track the coordinates of the human head and hand, respectively. Using the approach described in \cite{rakita2017motion}, we transform the position and orientation of the hand from the operator's frame to that of the manipulator. The RTDE library \cite{rtde_lib} is used to calculate the inverse kinematics of the UR3 and transmit the corresponding joint angles to the manipulator. 

To perform grasping tasks, the Robotiq 2F-85 gripper was used, for which a special control interface was developed to control the gripper. This device measures the angle between the thumb and the index finger and maps it to gripper position. It also serves as a mounting point for the HTC hand tracker and equipped with a button to switch between two cameras.

For robotic head control, we use the roll and pitch angles from the headset, as the robotic head has two degrees of freedom. During experiments, we encountered VR motion sickness, which occurs when there is a discrepancy between the visual information received through the VR headset and the head's physical movements, resulting in sensory conflict. To mitigate this issue, we switched to the UDP protocol and placed the head control in a separate thread with a 100 Hz control frequency. Unity \cite{unity} was used for video streaming to the headset, as well as for dataset collection.

\begin{figure}[htp]
    \centering
    \includegraphics[width=1\linewidth]{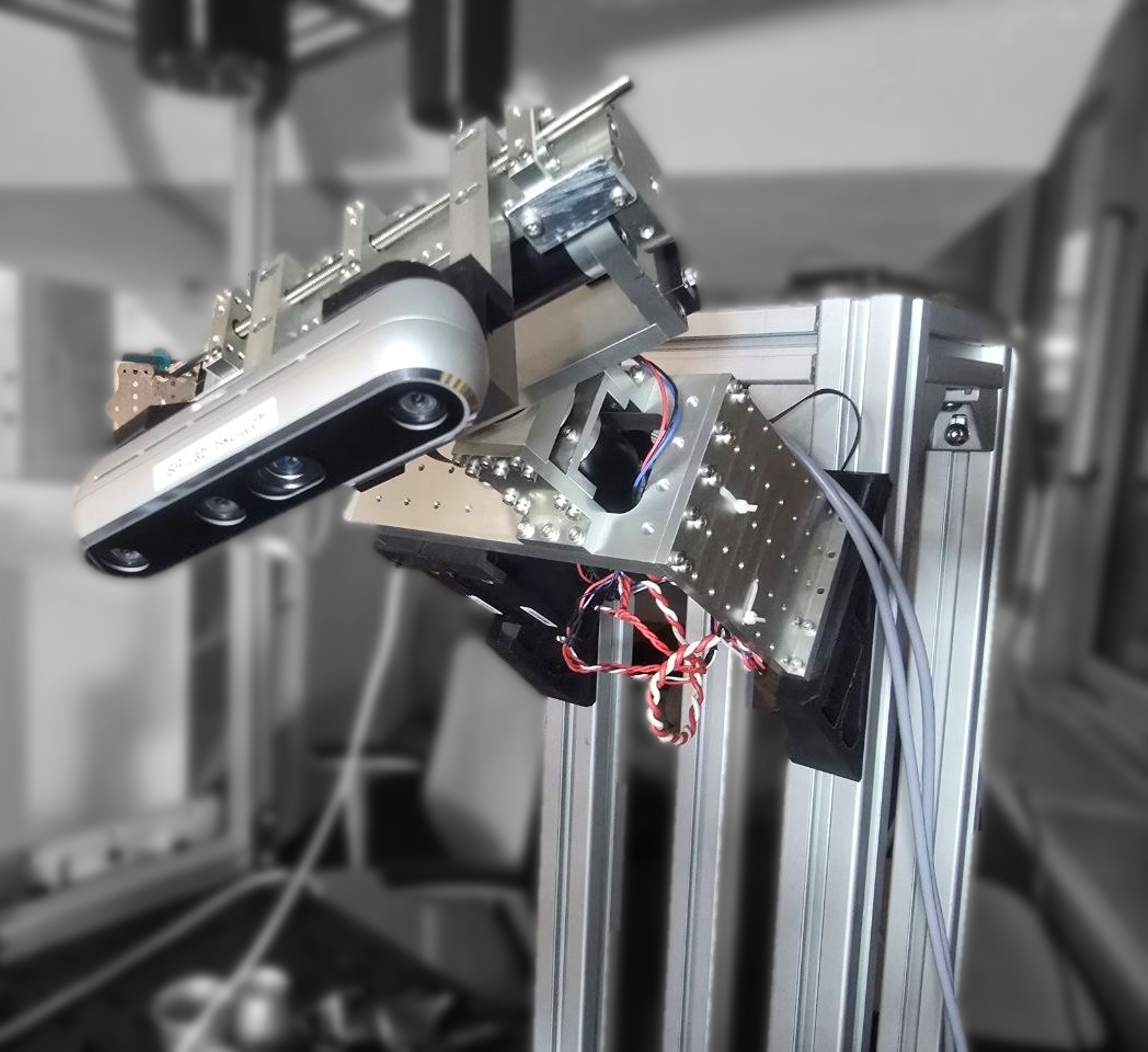}
    \caption{2-DoF Robotic head with Intel RealSense D455 Depth Camera.}
    \label{fig:head}
    \vspace{-0.3cm}
\end{figure}

On the robot side, a special robotic head was developed (\autoref{fig:head}) to help the operator maintain camera focus on the current manipulation object and provide a wider working area for the robot. We used a 2-DoF head powered by two Maxon DC motors. For low-level control, EPOS2 24/5 positioning controllers were employed. A homing procedure is required at each system startup due to the use of incremental encoders with the motors. Two Hall sensors were mounted on the head, and a driver was developed to connect them to the EPOS controllers. For high-level control and communication with the Operator side, we used an OrangePi 5 with ROS2 onboard. We adapted the ROS1 code from \cite{maxon_driver}, ported it to ROS2, and added features essential for our task. The Intel RealSense D455 Depth Camera mounted on the robotic head was connected directly to the operator's PC to reduce latency, though it can also be connected through the OrangePi if the Operator and Robot subsystems are remote.

\section{Experimental Evaluation} 

\subsection{Experimental Design, Tasks, \& Procedure}

\textit{Observation methods} — During the experiment, two approaches for workspace observation were compared. In both cases, two cameras were utilized: the Intel RealSense D455 and the HD Pro Webcam C920 USB webcam. The webcam was mounted on the gripper to provide a close-up view, while the RealSense camera was attached to the moving part of the robotic head. As a baseline, a fixed head was positioned to provide an optimal general view of the work area. In the second approach, the robotic head was allowed to move, enabling evaluation of the proposed teleoperated head system.

\textit{Subjects} — Nine participants, three women and six men, aged from 16 to 31 years (23.6 $\pm$4.09) took part in the experiment. The participants were informed about the experiment and filled out the consent form.

\begin{figure}[h!]
    \centering
    
    \begin{minipage}[b]{0.32\linewidth}
        \centering
        \includegraphics[width=\linewidth]{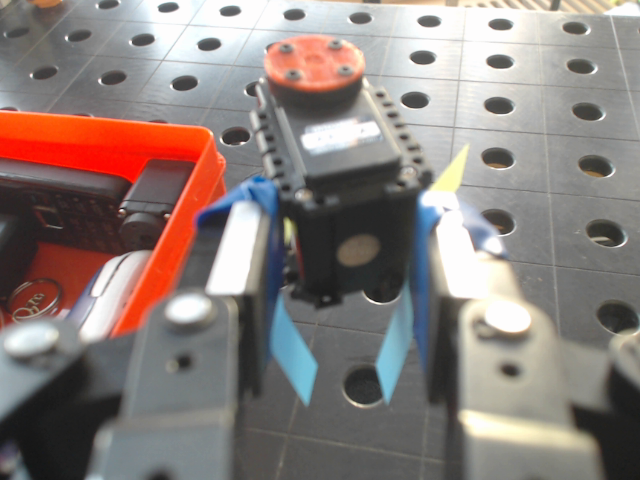}

    \end{minipage}
    \hfill
    \begin{minipage}[b]{0.32\linewidth}
        \centering
        \includegraphics[width=\linewidth]{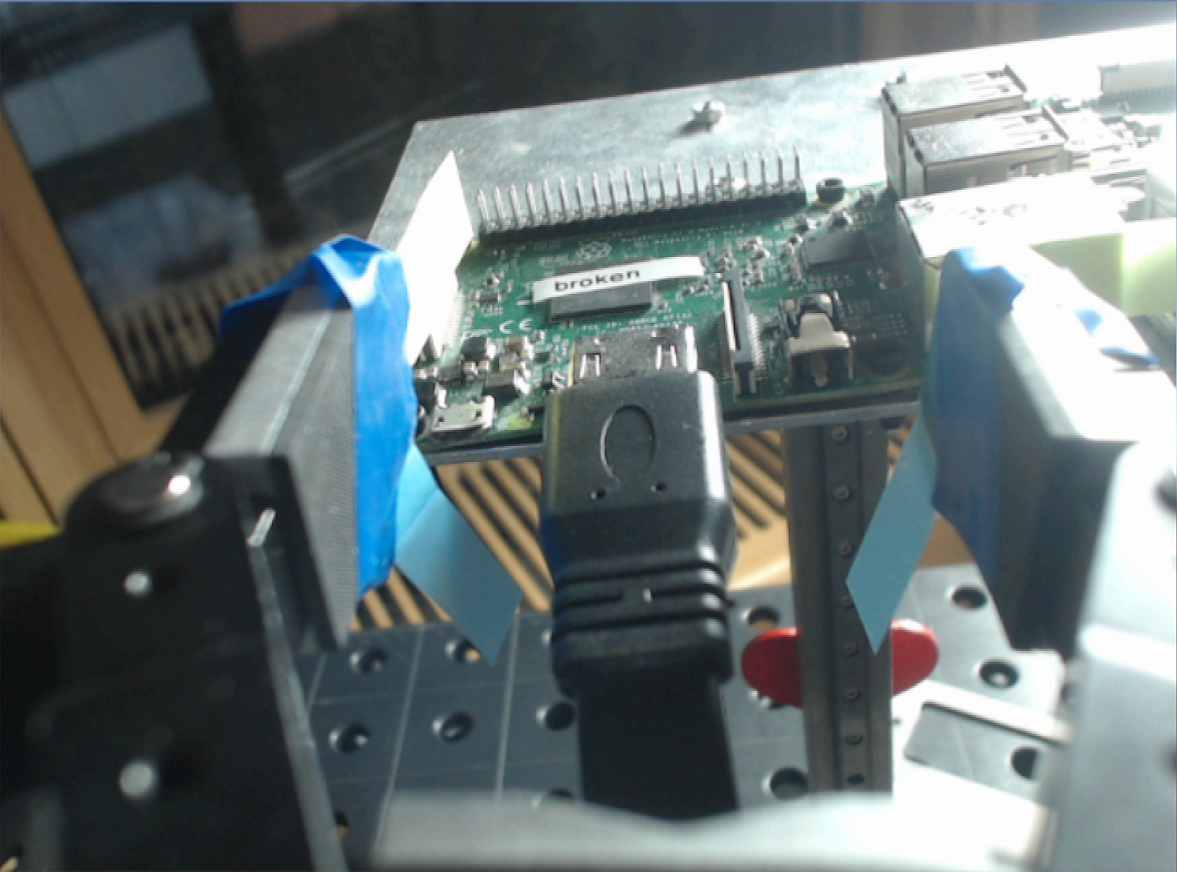}
    \end{minipage}
    \hfill
    \begin{minipage}[b]{0.32\linewidth}
        \centering
        \includegraphics[width=\linewidth]{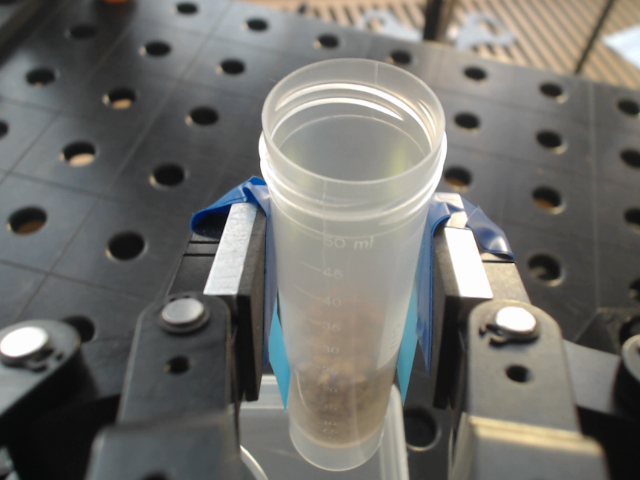}
    \end{minipage}

    \vspace{0.1cm} 
    
    \begin{minipage}[b]{0.32\linewidth}
        \centering
        \includegraphics[width=\linewidth]{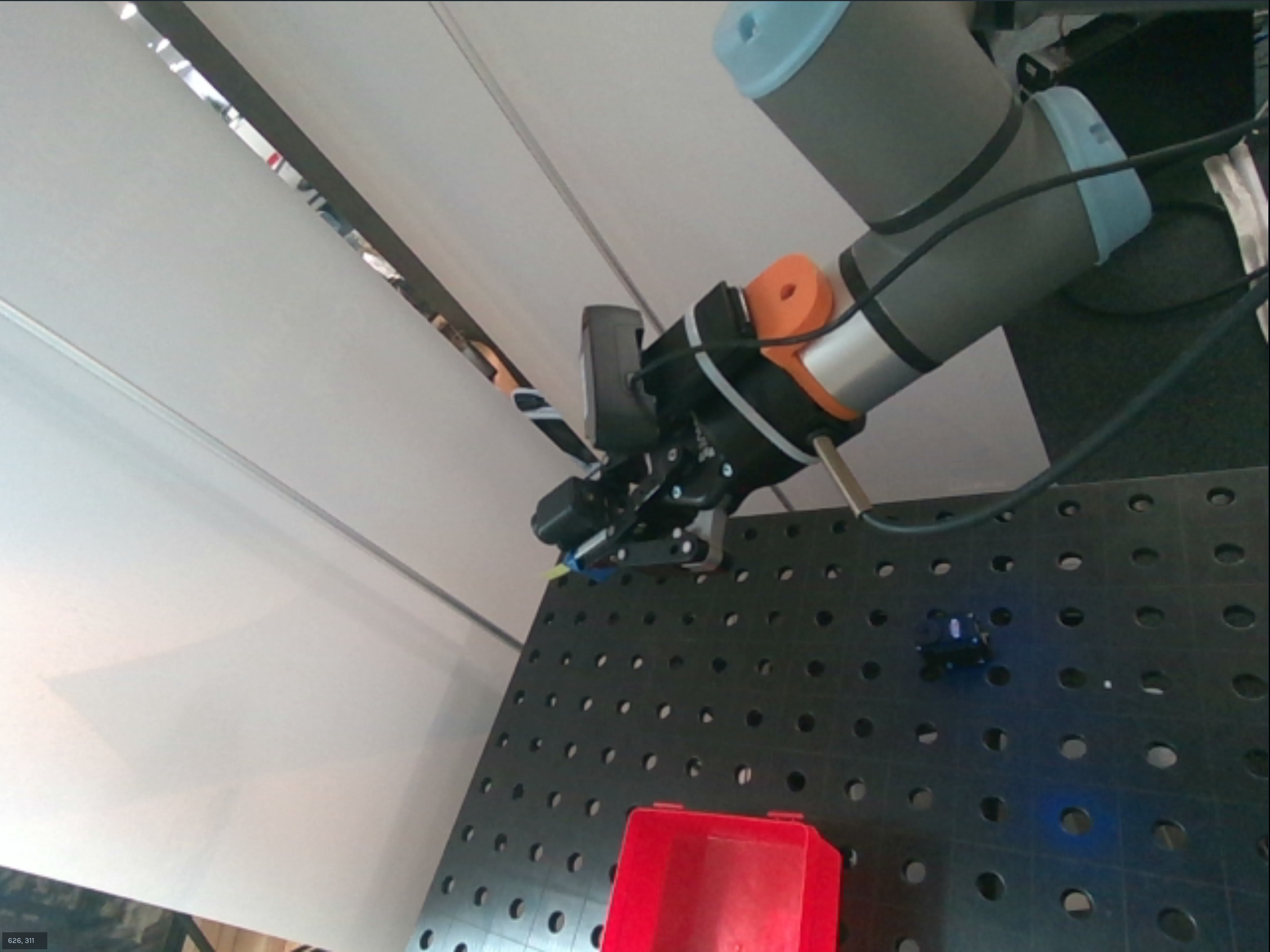}
    \end{minipage}
    \hfill
    \begin{minipage}[b]{0.32\linewidth}
        \centering
        \includegraphics[width=\linewidth]{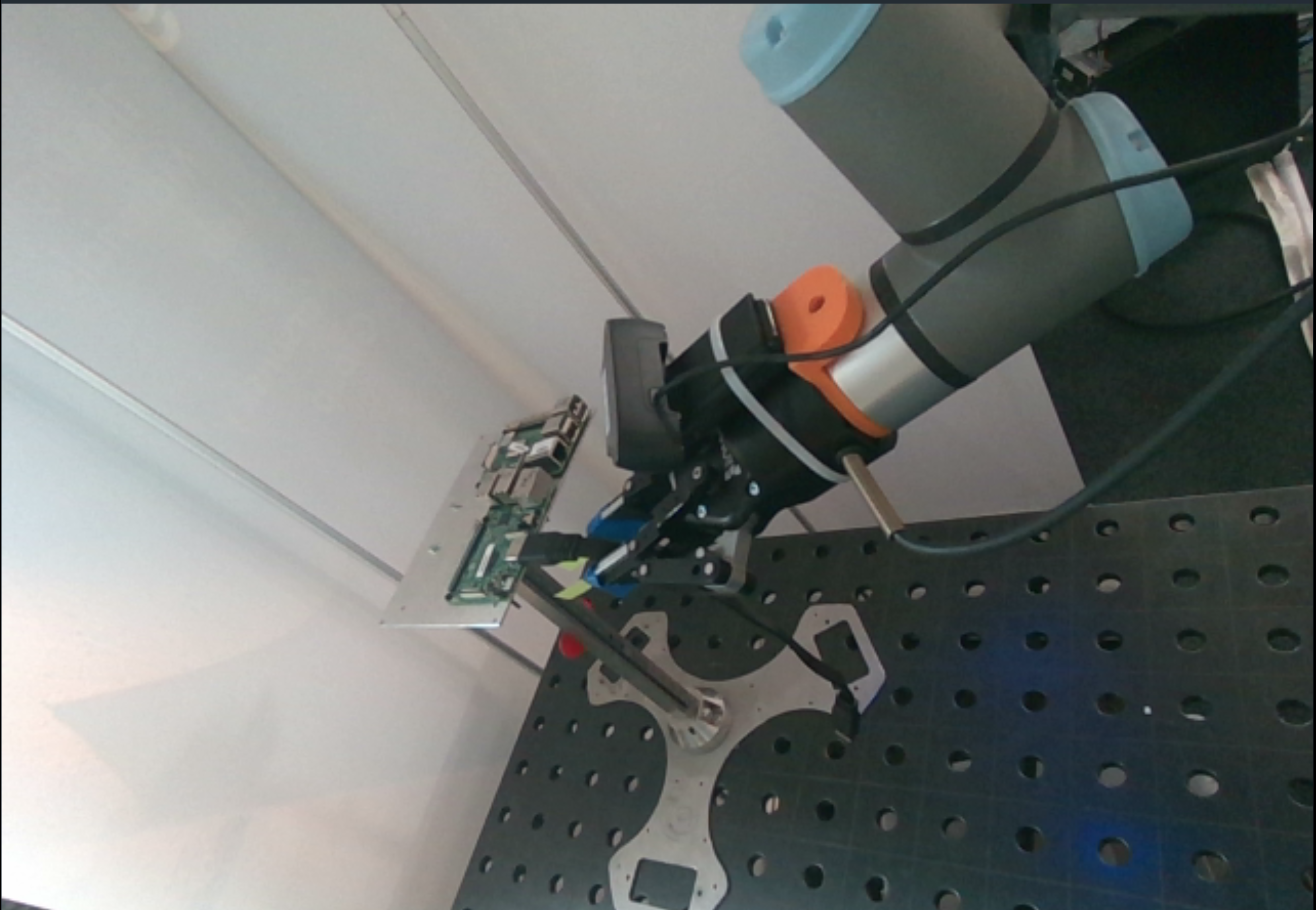}
    \end{minipage}
    \hfill
    \begin{minipage}[b]{0.32\linewidth}
        \centering
        \includegraphics[width=\linewidth]{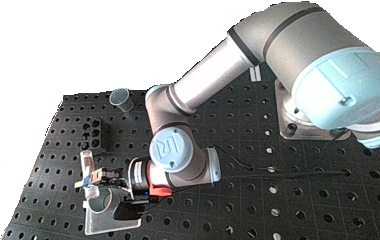}
    \end{minipage}

    \vspace{0.1cm} 
    
    \begin{minipage}[b]{0.32\linewidth}
        \centering
        \includegraphics[width=\linewidth]{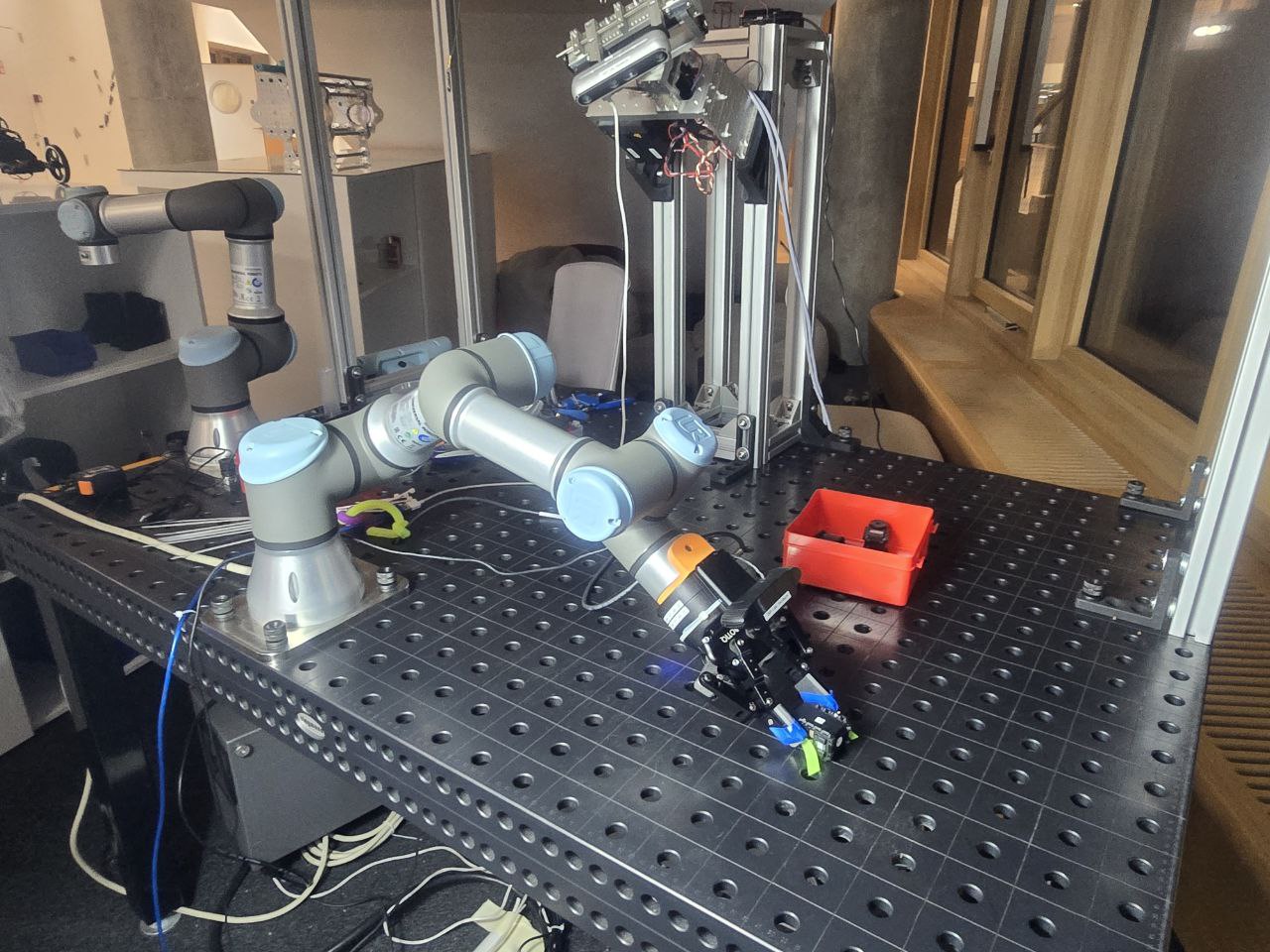}
        \caption*{``Pick\&place servo" \\  }
    \end{minipage}
    \hfill
    \begin{minipage}[b]{0.32\linewidth}
        \centering
        \includegraphics[width=\linewidth]{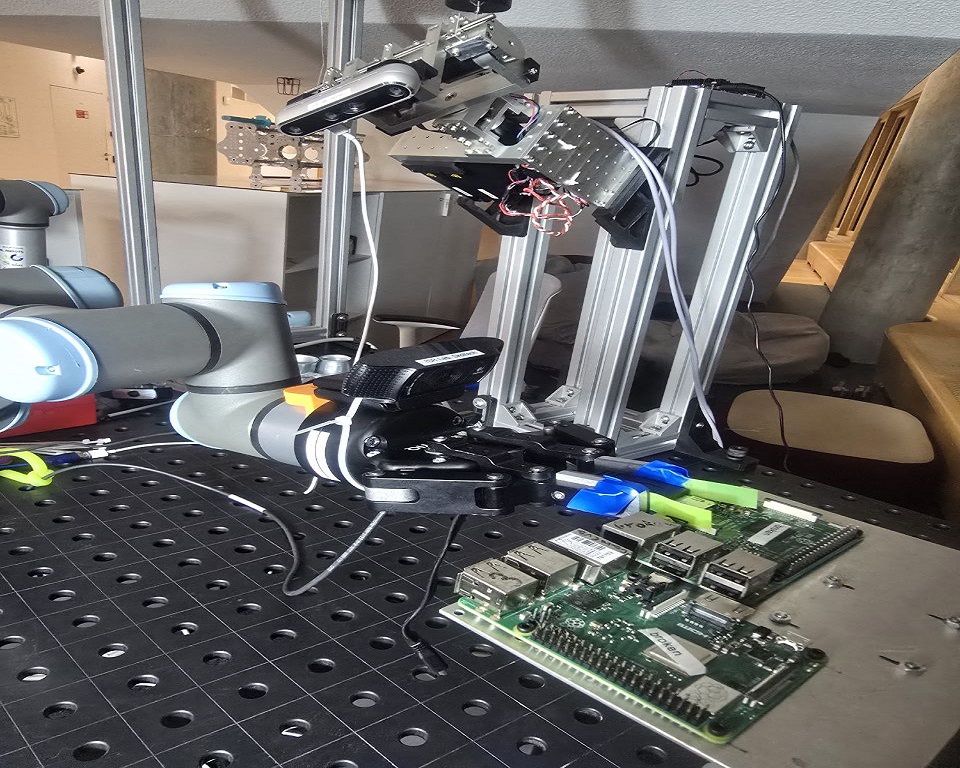}
        \caption*{``Removing USB"}
    \end{minipage}
    \hfill
    \begin{minipage}[b]{0.32\linewidth}
        \centering
        \includegraphics[width=\linewidth]{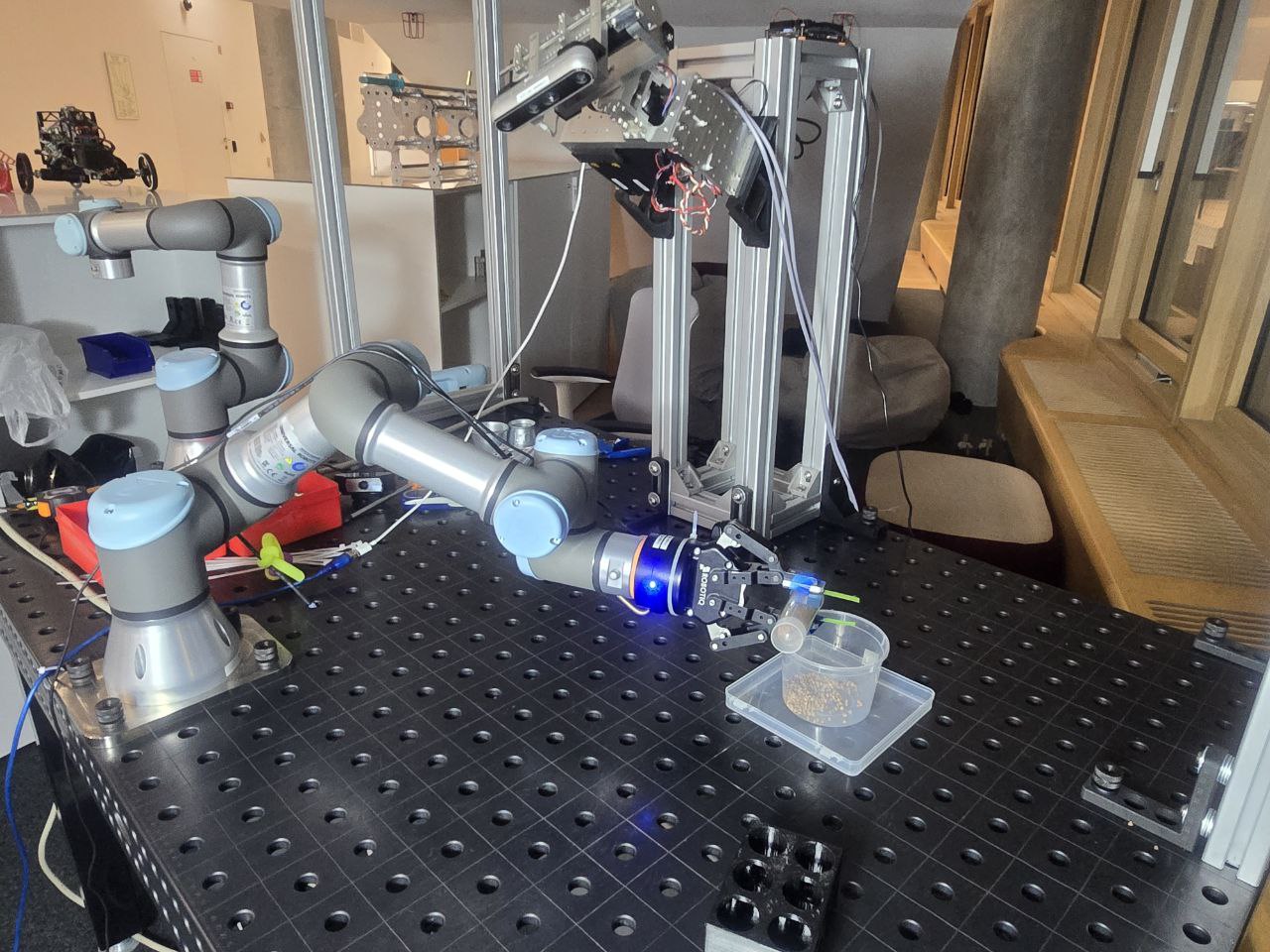}
        \caption*{``Pouring the cereal"}
    \end{minipage}

    \caption{Three tasks used in our user study. View from wrist (top row), head (middle row) and side (bottom row) cameras.}
    \label{fig:comparison}
\end{figure}

\textit{Tasks} — To ensure the applicability and generalizability of our teleoperation system for a wide range of physical manipulation tasks, we developed three tasks of varying difficulty levels (\autoref{fig:comparison}). The first task, a simple pick-and-place operation, required the participant to pick up a servo motor and place it in a box. We selected these motors due to their rectangular shape, which necessitates orienting the gripper fingers parallel to the edges of the rectangle. The second task involved removing a USB or HDMI connector from its socket, which tested manipulation accuracy, as the connector is small and applying force in the wrong direction could damage the socket. In the third task, the participant was asked to pick up a tube of cereal and pour it into another container. Cereal was chosen over water to prevent potential damage to the electronics.

\textit{Procedure} — After providing informed consent, participants were briefed on the study's objectives and task requirements. Before starting each observation method, participants received training on the control system and essential safety procedures. Once participants demonstrated readiness, they practiced a preliminary task: picking up and placing a plastic cup into a box. This task, selected for its simplicity, was designed to familiarize participants with the system. Each participant completed the training task at least three times, with the option to repeat it up to two additional times.

Following the training phase, participants completed the study tasks in sequence — pick-and-place servo, remove the connector, and pour the cereal — for each observation method. The tasks were ordered progressively to ensure a gradual increase in difficulty. The robot was reset to its initial configuration between tasks to maintain experimental consistency.

After completing the experiment, participants filled out a NASA Task Load Index (NASA-TLX) questionnaire \cite{NASA} to evaluate their subjective workload experience. NASA-TLX is a widely used workload assessment tool developed by NASA's Human Performance Group. It measures perceived workload across six subscales (each rated on a 1–20 scale): mental demand, physical demand, temporal demand, performance, effort and frustration.


\subsection{Experimental Results}
The results of the NASA-TLX questionnaire are summarized in Table~\ref{CM1}. The average success rate was 77.78\% for the dynamic head camera and 62.96\% for the static one. Despite the improvement in overall performance, higher levels of Mental Demand, Effort, and Frustration were reported with the dynamic head camera. This increase can be attributed to some participants experiencing motion sickness during the experiment. Although the communication delay was minimal, latency caused by motor inertia contributed to this issue.

It should also be noted that during the experiments, all items were placed within the field of view of the static camera. Consequently, the potential benefits of the increased field of view provided by the dynamic head camera were not evaluated.

\begin{table}[htbp]
\centering{
\caption{\textsc{NASA-TLX Average Rating for the Robot Telemanipulation Methods}}
\label{CM1}
\setlength{\tabcolsep}{8pt} 

\begin{tabular}{|l|l|l|l}
\cline{1-3}
                         & \textbf{Static head} & \textbf{Dynamic head}  &  \\ \cline{1-3}
\textbf{Mental Demand}   & 10.44          & 10.88                    &  \\ \cline{1-3}
\textbf{Physical Demand} & 10.88    & 9.88                       &  \\ \cline{1-3}
\textbf{Temporal Demand} & 5.11    & 6.67                     &  \\ \cline{1-3}
\textbf{Performance}     & 8.33    & 10.67                   &  \\ \cline{1-3}
\textbf{Effort}          & 13.88     & 14.22                     &  \\ \cline{1-3}
\textbf{Frustration}     & 5.33    & 5.66                   &  \\ \cline{1-3}
\end{tabular}}
\end{table}

During the experiments, it was found that the main challenge for most users was a lack of understanding of the robot's embodiment and the boundaries of its working area. Since the UR3 does not have built-in protection against self-collision, this was the most frequent cause of emergency stops, even though joint limits were implemented in the code. To address this issue, we updated our instructions to include a recommendation for operators to regularly verify the robot's configuration using the head-mounted camera.

Another significant issue was the absence of tactile feedback, which was particularly problematic during the connector removal task. Without the ability to feel resistance from the socket, operators often caused socket damage or triggered emergency stops. Additionally, some limitations of the gripper control interface became evident during this task. Since the thumb and index fingers are used both to control the gripper and to hold the control device, closing the gripper often caused unintentional displacement of the end-effector, reducing teleoperation accuracy.

In the third task, which was completed only by highly experienced participants, we encountered a lack of sufficient observational data. The wrist-mounted camera was positioned too close to effectively observe the target container, while the head camera's view was often partially obstructed by the robot's joints. To overcome this limitation, we plan to add a third camera to the setup. Additionally, some participants suggested that a third degree of freedom for the head, specifically for height adjustment, would improve the ability to assess the positions of manipulation objects.

\section{Conclusion and Future Work}

This study presents an immersive teleoperation system combining the HTC Vive VR setup and a 2-DoF robotic head to improve operator control and visual feedback during robotic manipulation. By enabling adaptive camera perspectives aligned with the operator’s head movements, the system enhances spatial awareness and increases task success rates by 15\% compared to camera setup with static head.


In the future, our aim is to train imitation learning policies on the dataset collected with the proposed teleoperation system.

\section*{Acknowledgment} 

Research reported in this publication was financially supported by the RSF grant No. 24-41-02039.


\bibliographystyle{IEEEtran}

\bibliography{sample-base} 

\end{document}